\documentclass[runningheads]{llncs}
\usepackage{graphicx}
\usepackage{makecell}
\usepackage{amsmath}
\usepackage{amsfonts,amssymb} 
\usepackage{caption}
\usepackage{subfigure}
\usepackage{wrapfig}

\usepackage{parskip}
\usepackage{booktabs}
\usepackage{multirow}
\usepackage{xcolor}

\begin{document}
\title{ViT-DAE: Transformer-driven Diffusion Autoencoder for Histopathology Image Analysis}
%

\author{Xuan Xu \and
Saarthak Kapse \and
Rajarsi Gupta \and
Prateek Prasanna}

\authorrunning{F. Author et al.}

\institute{Stony Brook University, NY, USA \\
\email{xuaxu@cs.stonybrook.edu}\\
\email{\{saarthak.kapse, prateek.prasanna\}@stonybrook.edu}\\
\email{Rajarsi.Gupta@stonybrookmedicine.edu}}

%
%

\maketitle              
\begin{abstract}

Generative AI has received substantial attention in recent years due to its ability to synthesize data that closely resembles the original data source. While Generative Adversarial Networks (GANs) have provided innovative approaches for histopathological image analysis, they suffer from limitations such as mode collapse and overfitting in discriminator. Recently, Denoising Diffusion models have demonstrated promising results in computer vision. These models exhibit superior stability during training, better distribution coverage, and produce high-quality diverse images. Additionally, they display a high degree of resilience to noise and perturbations, making them well-suited for use in digital pathology, where images commonly contain artifacts and exhibit significant variations in staining. In this paper, we present a novel approach, namely ViT-DAE, which integrates vision transformers (ViT) and diffusion autoencoders for high-quality histopathology image synthesis. This marks the first time that ViT has been introduced to diffusion autoencoders in computational pathology, allowing the model to better capture the complex and intricate details of histopathology images. We demonstrate the effectiveness of ViT-DAE on three publicly available datasets. Our approach outperforms recent GAN-based and vanilla DAE methods in generating realistic images.

\end{abstract}
\section{Introduction}

Over the last few years, generative models have sparked significant interest in digital pathology~\cite{jose2021generative}. The objective of generative modeling techniques is to create synthetic data that closely resembles the original or desired data distribution. The synthesized data can improve the performance of various downstream tasks, eliminating the need for obtaining large-scale costly expert annotated data. Methods built on Generative Adversarial Networks (GANs) have provided novel approaches to address various challenging histopathological image analysis problems including stain normalization~\cite{runz2021normalization}, artifact removal~\cite{dahan2022artifact}, representation learning~\cite{boyd2021self}, data augmentation~\cite{wei2019generative,xue2021selective}, etc.
However, GAN models experience mode collapse and limited latent size~\cite{dae}. 
Their discriminator is prone to overfitting while producing samples from datasets with imbalanced classes~\cite{xiao2021tackling}. This results in lower quality image synthesis.

Diffusion models, on the other hand, are capable of producing images that are more diverse. They are also less prone to overfitting compared to GANs~\cite{xiao2021tackling}. Recently, Denoising Diffusion Probabilistic models (DDPMs)~\cite{ho2020denoising} and score-based generative models~\cite{song2020score} have shown promise in computer vision. Ho et al. present high quality image synthesis using diffusion probabilistic models~\cite{ho2020denoising}. Latent diffusion models have been proposed for high-resolution image synthesis which are widely applied in super resolution, image inpainting, and semantic scene synthesis~\cite{rombach2022high}.  These models have been shown to achieve better synthetic image quality compared to GANs~\cite{dhariwal2021diffusion}. Denoising Diffusion Implicit models (DDIMs) construct a class of non-Markovian diffusion processes which makes sampling from reverse process much faster~\cite{song2020denoising}. This modification in the forward process preserves the goal of DDPM and allows for deterministically encoding an image to the noise map. Unlike DDPMs~\cite{moghadam2023morphology}, DDIMs enable control over image synthesis owing to the latent space flexibility (attribute manipulation)~\cite{dae}.  

\textbf{Motivation:} Diffusion models have the potential to make a significant impact in computational pathology. Compared with other generative models, diffusion models are generally more stable during training. Unlike natural images, histopathology images are intricate and harbor rich contextual information which can make GANs difficult to train and suffer from issues like mode collapse where the generator produces limited and repetitive outputs failing to capture the complex diversity in histopathology image distribution.
Diffusion models, on the other hand, are more stable and provide better distribution coverage leading to high fidelity diverse images. They are considerably more resistant to perturbations and noise than GANs, which is crucial in digital pathology because images routinely contain artifacts and exhibit large variations in staining~\cite{kanwal2022devil}. 
With desirable attributions in high quality image generation, diffusion models have the capacity to enhance various endeavors in computational pathology including image classification, data augmentation, and super resolution. Despite their potential benefits, diffusion models remain largely unexplored in computational pathology.

Recently, Preechakul et al.~\cite{dae} proposed a diffusion autoencoder (DAE) framework which encodes natural images into a representation using semantic encoder and uses the resulting semantic subcode as the condition in the DDIM image decoder. The encoding of histopathology images, however, presents a significant challenge, primarily because these images contain intricate microenvironments of tissues and cells, which have complex and diverse spatial arrangements. Thus, it is imperative to use a semantic encoder that has a high capacity to represent and understand the complex and global spatial structures present in histopathology images. Towards this direction, we propose to introduce vision transformer (ViT)~\cite{dosovitskiy2020image} as the semantic encoder instead of their convolutional neural network (CNN) counterpart in DAE; our proposed method is called ViT-DAE. The self-attention~\cite{vaswani2017attention} mechanism in ViT allows to better capture global contextual information through long-range interactions between the regions/patches of images. A previous study~\cite{naseer2021intriguing} has shown that 1) ViTs perform better than CNNs and are comparable to humans on
shape recognition, 2) ViTs are more robust against perturbations and image corruptions. Based on these findings, ViT presents a promising solution in encoding meaningful and rich representations of complex and noisy tissue structures, where shape and spatial arrangement of biological entities form two crucial motifs. Hence we hypothesize that a transformer-based semantic encoder in DAE would result in higher quality histopathology image synthesis. 

To summarize our main contributions, \textbf{(1)}
We are the first to introduce conditional DDIM in histopathology image analysis, and \textbf{(2)} We enhance the conditional DDIM by incorporating ViT as a semantic encoder, enabling it to holistically encode complex phenotypic layout specific to histopathology. We demonstrate the effectiveness of ViT-DAE on three public datasets; it outperforms recent GAN-based and vanilla DAE methods, in generating better images.


\begin{figure}[t]
\begin{center}
\includegraphics[width=0.8\linewidth]{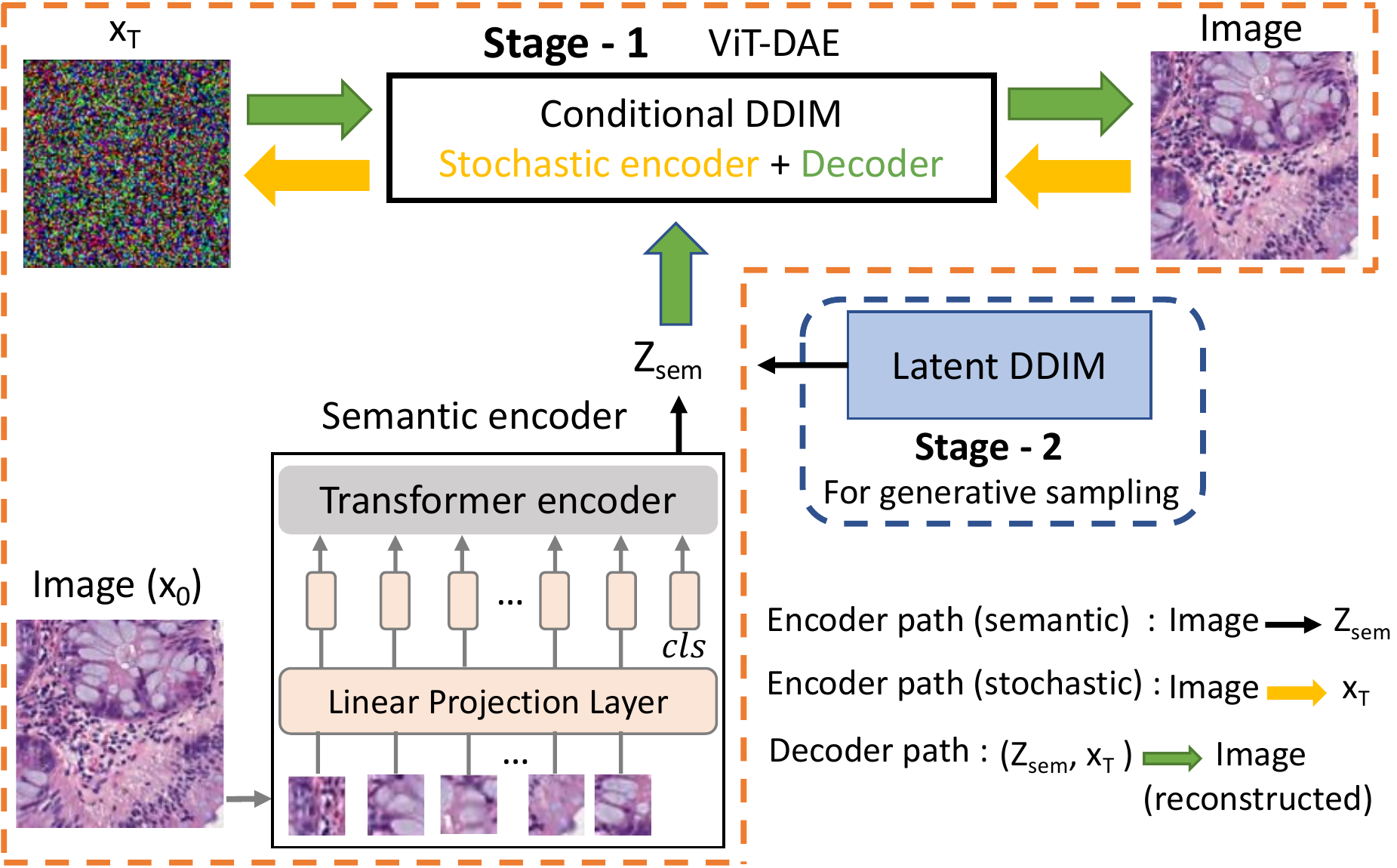}
\end{center}
\caption{Overview of the proposed ViT-DAE framework. \textbf{Training:} In Stage-1, an input image is encoded into a semantic representation by the ViT. 
This representation is taken as the condition for the conditional DDIM to decode the noisy image. In Stage-2, a latent DDIM is trained to learn the distribution of semantic representations of data. \textbf{Generative sampling:} We synthesize the semantic representations from the latent DDIM and feed it to the conditional DDIM along with randomly initialized noisy image to generate new histopathology samples.}\label{overview}
\vspace{-0.4cm}
\end{figure}

\section{Proposed Method}
To generate histopathology images that are meaningful and of diagnostic quality, we propose ViT-DAE, a framework that utilizes a transformer-enhanced diffusion autoencoder. Our method consists of two stages of \textbf{training}. \textit{Stage-1} comprises i) a ViT-based semantic encoder, which captures the global semantic information of an \textit{input image}, and ii) a conditional DDIM which is an autoencoder, takes in input the semantic representation by ViT as a condition and the noisy image to reconstruct the \textit{input image}. In \textit{Stage-2} with the frozen semantic encoder, a latent DDIM is trained to learn the distribution of semantic representation of the data in the latent space. Following this, for \textbf{Generative sampling}, first the latent DDIM is fed with a noisy vector, outputting a synthesized sample from the learned semantic representation distribution. This, along with a randomly initialized noisy image is then fed to the conditional DDIM for image generation. An overview of the proposed framework is shown in Fig.~\ref{overview}.



\textbf{Vision-transformer enhanced semantic encoder.}
An input image $\rm{\mathbf{x}_0}$ is encoded into a semantic representation $\rm{\mathbf{z}_{sem}}$ via our ViT based semantic encoder $\rm{\mathbf{z}_{sem}=Enc_{\phi}(\mathbf{x}_0)}$. We split the input image into patches and apply a linear projection followed by interacting them in the transformer encoder. The output class token, $cls$, is projected to dimension $d=512$ via a linear layer, and then is used as a condition for the decoder part of DAE. The semantic representations from the input images encoded by the ViT provide an information-rich latent space which is then utilized as the condition in DDIM following~\cite{dae}.


\textbf{Conditional DDIM.}
A Gaussian diffusion process can be described as the gradual addition of small amount of Gaussian noise to input images in $T$ steps which leads to a sequence of noisy images $\mathbf{x}_1,...,\mathbf{x}_T$~\cite{ho2020denoising,dae}. 
At a given time $t$ (out of $T$), the diffusion process can be defined as $q(\mathbf{x}_t|\mathbf{x}_{t-1})=\mathcal{N}(\sqrt{1-\beta_t}\mathbf{x}_{t-1},\beta_t\mathbf{I})$, where $\beta_t$ are the noise level hyperparameters. The corresponding noisy image of $\mathbf{x}_0$ at time $t$ is another Gaussian $q(\mathbf{x}_t|\mathbf{x}_0)=\mathcal{N}(\sqrt{\alpha_t}\mathbf{x}_0,(1-\alpha_t)\mathbf{I})$ where $\alpha_t=\prod_{s=1}^t(1-\beta_s)$. This is followed by learning of the generative reverse process, i.e., the distribution $p(\mathbf{x}_{t-1}|\mathbf{x}_t)$~\cite{ho2020denoising,dae}.
DDIM~\cite{song2020denoising} proposes this reverse process as a deterministic generative process, given by:
\begin{equation}
    \mathbf{x}_{t-1}=\sqrt{\alpha_{t-1}}\Bigg(\frac{\mathbf{x}_t-\sqrt{1-\alpha_t}\epsilon_\theta^t(\mathbf{x}_t)}{\sqrt{\alpha_t}}\Bigg)+\sqrt{1-\alpha_{t-1}}\epsilon_\theta^t(\mathbf{x}_t)\label{eq1}
\end{equation}
where $\epsilon_\theta^t(\mathbf{x}_t)$, proposed by~\cite{ho2020denoising}, is a function which takes the noisy image $\mathbf{x}_t$ and predicts the noise using a UNet~\cite{ronneberger2015u}. The inference distribution is given by:
\begin{equation}
    q(\mathbf{x}_{t-1}|\mathbf{x}_t,\mathbf{x}_0)=\mathcal{N}\Bigg(\sqrt{\alpha_{t-1}}\mathbf{x}_0+\sqrt{1-\alpha_{t-1}}\frac{\mathbf{x}_t-\sqrt{\alpha_t}\mathbf{x}_0}{\sqrt{1-\alpha_t}},\mathbf{0}\Bigg)\label{eq2}
\end{equation}

The conditional DDIM decoder takes an input in the form of $\mathbf{z}=(\mathbf{z}_{\rm{sem}},\mathbf{x}_T)$ to generate the output images. Conditional DDIM leverages the reverse process defined in 
 Eqn. \ref{eq3}, \ref{eq4} to model 
$p_\theta(\mathbf{x}_{t-1}|\mathbf{x}_t,\rm\mathbf{z}_{sem})$ to match the inference distribution $q(\mathbf{x}_{t-1}|\mathbf{x}_t,\mathbf{x}_0)$ defined in Eqn. \ref{eq2}.

\begin{equation}  
	p_\theta(\mathbf{x}_{0:T}|\mathbf{z}_{\rm{sem}})=
        p(\mathbf{x}_T)\prod \limits_{t=1}^Tp_{\theta}(\mathbf{x}_{t-1}|\mathbf{x}_{t},\rm{\mathbf{z}_{sem}}) \label{eq3}
\end{equation}

\begin{equation}
p_{\theta}(\mathbf{x}_{t-1}|\mathbf{x}_{t},\rm{\mathbf{z}_{sem}})=
\begin{cases}
\mathcal{N}(\mathbf{f}_\theta(\mathbf{x}_1,1,\mathbf{z}_{\rm{sem}}),\mathbf{0})& \text{if}\ t=1\\
q(\mathbf{x}_{t-1}|\mathbf{x}_t,\mathbf{f}_\theta(\mathbf{x}_t,t,\mathbf{z}_{\rm{sem}})) &\text{otherwise}
\end{cases} \label{eq4}
\end{equation}

where $\mathbf{f}_\theta$ in Eqn. \ref{eq4} is parameterized as the noise prediction network $\epsilon_\theta(\mathbf{x}_t,t,\mathbf{z}_{\rm{sem}})$ from Song et al.~\cite{song2020denoising}:
\begin{equation}
    \mathbf{f}_\theta(\mathbf{x}_t,t,\mathbf{z}_{\rm{sem}})=\frac{1}{\sqrt{\alpha_t}}(\mathbf{x}_t-\sqrt{1-\alpha_t}\epsilon_\theta(\mathbf{x}_t,t,\mathbf{z}_{sem}))
\end{equation}
This network is a modified version of a UNet from~\cite{dhariwal2021diffusion}.


\textbf{Generative sampling.}
To generate images from the diffusion autoencoder, a latent DDIM is leveraged to learn the semantic representation distribution of $\mathbf{z}_{\rm{sem}}= \rm{Enc_\phi(x_0)}$, $\mathbf{x}_0\sim p(\mathbf{x}_0)$. We follow the framework from~\cite{dae} to leverage the deep MLPs (10-20 layers) with skip connections as the latent DDIM network. Loss $\mathcal{L}_{\rm{latent}}$ is optimized during training with respect to latent DDIM's parameter, $\omega$:
\begin{equation}
    \mathcal{L}_{\rm{latent}}=\sum \limits_{t=1}^T\mathbb{E}_{\mathbf{z}_{\rm{sem}},\epsilon_t}\Big [||\epsilon_\omega(\mathbf{z}_{\rm{sem},t},t)-\epsilon_t||_1\Big]\label{eq9}
\end{equation}
where $\epsilon_t\in\mathbb{R}^d\sim\mathcal{N}(\mathbf{0,I)}$, $\mathbf{z}_{\rm{sem},t}=\sqrt{\alpha_t}\mathbf{z}_{\rm{sem}}+\sqrt{1-\alpha_t}\epsilon_t$ and $T$ is the same as our conditional image decoder.

The semantic representations are normalized to zero mean and unit variance before being fed to the latent DDIM, to model the semantic representation distribution. Generative sampling using diffusion autoencoders involves three steps. First, we sample the $\mathbf{z}_{\rm{sem}}$ from the latent DDIM which learns the distribution of semantic representations and unnormalizes it. Then we sample $\mathbf{x}_T\sim\mathcal{N}(\mathbf{0,I})$. Finally we decode $\mathbf{z}=(\mathbf{z}_{\rm{sem}},\mathbf{x}_T)$ via the conditional DDIM image decoder. Note that for generating class-specific images, an independent latent DDIM model is trained on semantic distribution for each class. Whereas for class-agnostic sampling, just one latent DDIM is trained on the complete cohort.

\section{Experiments and Results}
\subsection{Datasets and Implementation Details} 

In this study, we utilized 4 datasets. The self-supervised (SSL) pretraining of the semantic encoder - vision transformer (ViT)~\cite{dosovitskiy2020image} using DINO~\cite{dino} framework is carried out on TCGA-CRC-DX~\cite{kather2019deep}, NCT-CRC-HE-100K~\cite{kather2018100}, and PCam~\cite{veeling2018rotation}. The corresponding pre-trained ViT models are then used as semantic encoders for training separate diffusion autoencoders~\cite{dae} on Chaoyang~\cite{zhu2021hard}, NCT-CRC-HE-100K~\cite{kather2018100}, and PCam~\cite{veeling2018rotation} datasets, respectively. Since the Chaoyang dataset contains only a few thousand images, DINO pre-training is conducted on another dataset (TCGA-CRC-DX) consisting of images from the same organ (colon). For NCT-CRC and PCam, the official train split provided for each dataset is used for self-supervision and diffusion autoencoder training. 
\textbf{TCGA-CRC-DX}~\cite{kather2019deep} consists of images of tumor tissue of colorectal cancer (CRC) WSIs in the TCGA database (N=368505). All the images are of size $512\times512$ pixels (px). This dataset is just utilized for self-supervised pre-training.
\textbf{Chaoyang}~\cite{zhu2021hard} contains a total of 6160 colon cancer images of size $512\times512$ px. These images are assigned one of the four classes - normal, serrated, adenocarcinoma, and adenoma by the consensus of three pathologists. The official train split is used to train the diffusion autoencoder as well as patch classification model; the official test split~\cite{zhu2021hard} is used to report the downstream classification performance.
\textbf{NCT-CRC-HE-100K}~\cite{kather2018100} contains 100k ($224\times224$ px) non-overlapping images from histological images of human colorectal cancer (CRC) and normal (H\&E)-stained tissue. There are nine classes in this dataset. 
\textbf{PCam}~\cite{veeling2018rotation} includes 327,680 images ($96\times96$ px) taken from histopathologic scans of lymph node sections in breast. Each image has a binary annotation indicating the presence of metastatic tissue. 

\textbf{Environment:}
Our framework is built in PyTorch 1.8.1 and trained on two Quadro RTX 8000 GPUs. We use ViT-Small as our transformer encoder. It is pretrained via DINO using default parameters~\cite{dino}. All images are resized to $224\times224$ px for SSL. In contrast, due to memory constraints for generative model training, the images are scaled to $128\times128$ px. To optimize the diffusion autoencoder, we adopted default parameters and configurations from DAE~\cite{dae}\\
\textbf{Evaluation metrics:}
We employ Frechet inception distance (\textbf{FID}), Improved Precision, and Improved Recall to evaluate the similarity between the distribution of synthesized images and real images~\cite{heusel2017gans,kynkaanniemi2019improved}. For FID computation~\cite{Seitzer2020FID}, the real and synthesized images are fed into an Inception V3 model~\cite{kynkaanniemi2022role} to extract features from pool\_3 layer. The FID method then calculates the difference between mean and standard deviation from these features.
A lower FID score indicates a higher similarity between the distributions. For NCT-CRC and PCam, we computed FID scores between all the real images from training set and our 50k generated images. For Chaoyang, FID is computed between all the training images and the generated 3k images.
Improved Precision (\textbf{IP}) and Recall (\textbf{IR}) estimate the distribution of real images and synthesized images by forming explicit, non-parameteric representations of the manifolds~\cite{kynkaanniemi2019improved}. IP describes the probability that a random generated image falls within the support of the real image manifold. Conversely, IR is defined as the probability that a random real image belongs to the generated image manifold.\\
For the downstream classification analysis with generated samples by ViT-DAE on Chaoyang dataset, we report accuracy as well as class-wise F1 score. \\
\textbf{Manifold visualization:} Motivated by~\cite{kynkaanniemi2019improved}, we generate the manifold. We employ PCA to reduce the dimensionality of the representation space; the top two PCs represent the transformed 2D feature space. We compute the radii for each feature vector by fitting the manifold algorithm to the transformed space. We then generate a manifold by plotting circles with the 2D vector as their centers and their corresponding radii as radius. This provides a comprehensive visualization and allows for an intuitive understanding of the learned representations.
\subsection{Results}
We compared two contemporary methods, Diffusion Autoencoder (DAE)~\cite{dae} and VQ-GAN~\cite{vqgan}, which are built on the latest advances in diffusion and GAN-based algorithms, respectively. Table~\ref{table:result:accuracy} compares the quality of synthesized images produced by ViT-DAE with other methods.

\begin{figure}[t]
\begin{center}
\includegraphics[width=.77\linewidth]{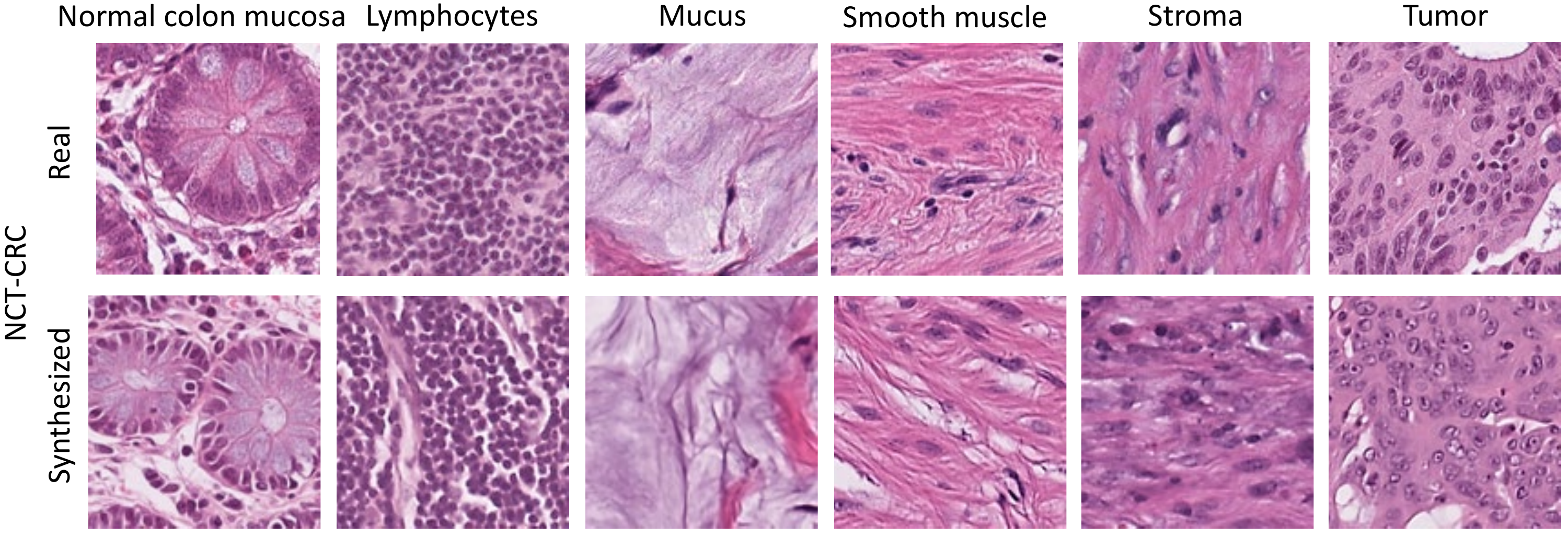}
\end{center}
\caption{Real and synthesized images using ViT-DAE on NCT-CRC} \label{visualization}
\vspace{-0.2cm}
\end{figure}

\textbf{Quantitative analysis:} Images produced by ViT-DAE have the lowest FID scores and highest IP and IR for the NCT-CRC and PCam datasets; the results are comparable with the other two approaches, particularly DAE, on the Chaoyang dataset. This consistent improvement may be attributed to ViT's replacement of the convolution-based semantic encoder. ViT has a far greater capacity to learn the contextual information and long-range relationships in the images than its CNN counterparts. As a result, a ViT-based semantic encoder could more effectively capture high level semantic information and provide more meaningful representation as the condition of DDIM, which improves the quality of generated images. CNN-based conditional DDIM faces the difficulty of encoding the complex spatial layouts in histopathology images. 
Since the CNN-based semantic encoder consists of 24.6M parameters compared to 21.6M in ViT-based encoder, we attribute this improvement to the superior global modeling of ViTs. 
\begin{table}[t]
\tiny
\caption{Comparison of FID, IP, IR on three datasets.}
\label{table:result:accuracy}
\begin{center}
\setlength{\tabcolsep}{1.0mm}{
\begin{tabular}{l |c c c | c c c | c c c}
\toprule
 \multicolumn{1}{c|}{Dataset} 
    & \multicolumn{3}{c|}{NCT-CRC} 
        & \multicolumn{3}{c|}{PCam}
            & \multicolumn{3}{c}{Chaoyang}
    
\\
\multicolumn{1}{c|}{Metric} 
    & \multicolumn{1}{c}{FID} $\downarrow$ & \multicolumn{1}{c}{IP} $\uparrow$ & \multicolumn{1}{c}{IR} $\uparrow$
        & \multicolumn{1}{c}{FID} $\downarrow$ & \multicolumn{1}{c}{IP} $\uparrow$ & \multicolumn{1}{c}{IR} $\uparrow$
            & \multicolumn{1}{c}{FID} $\downarrow$ & \multicolumn{1}{c}{IP} $\uparrow$ & \multicolumn{1}{c}{IR} $\uparrow$
    
\\
\midrule

\hfil VQ-GAN~\cite{vqgan}
    & $27.86$          & $0.57$   & $0.26$       
        & $15.99$          & $0.56$   & $0.22$  
            & $51.35$        & $0.43$   & $\textbf{0.53}$  
    
\\
\hfil DAE~\cite{dae}
    & $14.91$  & $0.58$   & $0.30$  
        & $39.42$  & $0.32$   & $0.28$  
            & $\textbf{35.69}$  & $0.50$   & $0.44$  

\\ 

\midrule
\hfil ViT-DAE (ours)
    & $\textbf{12.14}$  & $\textbf{0.60}$   & $\textbf{0.40}$  
        & $\textbf{13.39}$   & $\textbf{0.60}$   & $\textbf{0.44}$  
            & $36.18$   & $\textbf{0.51}$   & $0.50$  
    
\\
\bottomrule
\end{tabular}
}
\vspace{-0.1cm}
\end{center}
\end{table}
\textbf{Qualitative analysis:} 
 \begin{figure}[t]
 \centering
\includegraphics[width=0.82\textwidth]{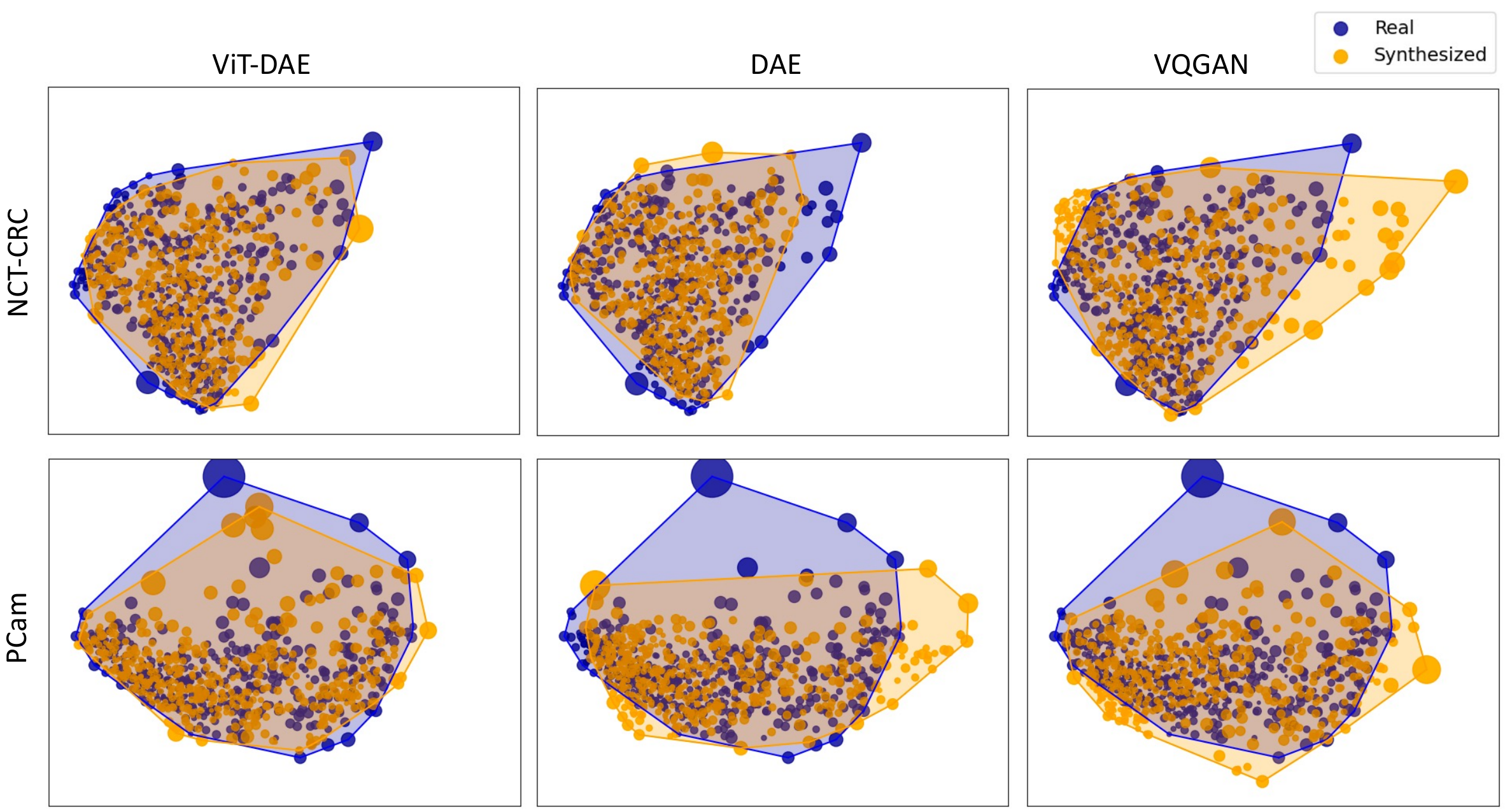}

\caption{Manifold visualization. Higher overlap indicates greater similarity.} 
\label{manifold}
\vspace{-0.2cm}
\end{figure}
 The similarity between distribution coverage of real and generated images is visually evaluated using manifold visualization in Fig.~\ref{manifold}. For the NCT-CRC and PCam datasets, our synthesized images have a closer distribution to the real image manifold compared to VQ-GAN and DAE. For Chaoyang (in supplementary), DAE and ViT-DAE generate distributions that are very comparable to real images while outperforming the generated manifold from VQ-GAN. We also provide class-wise samples along with pathologist's interpretations of synthesized images generated by ViT-DAE in Fig.~\ref{visualization}. We can see that our method captures the distribution sufficiently well and generates reasonably plausible class-specific images (more in supplementary). Here we summarize our pathologist's impressions on ``\textit{How phenotypically real are the synthesized images?}" for the different classes in NCT-CRC. \textit{Normal mucosa:} Realistic in terms of cells configured as glands as the main structural element of colonic mucosa, location of nuclei at the base of glands with apical mucin adjacent to the central lumen.
\textit{Lymphocytes:} Realistic in terms of cell contours, sizes, color, texture, and distribution. \textit{Mucus:} Color and texture realistic for mucin (mucoid material secreted from glands). \textit{Smooth Muscle:} Realistic with central nuclei in elongated spindle cells with elastic collagen fibers and no striations. \textit{Cancer associated stroma:} Realistic in terms of reactive stroma with inflammatory infiltrate (scattered lymphs, neutrophil) and increased cellularity of stromal cells. \textit{Tumor:} Realistic with disordered growth via nuclear crowding with a diversity of larger than normal epithelial (glandular epithelial) cells and irregular shaped cells. \\
\textbf{Downstream analysis:} To assess the efficacy of synthesized images, we conducted a proof-of-concept investigation by designing a  classification task on the Chaoyang dataset. We trained a classification model exclusively on class-conditioned synthesized images and evaluated its performance on real images from official test split~\cite{zhu2021hard}. The performance was comparable to that of a classification model trained solely on real images from provided train split. 
Our results demonstrate that a hybrid training approach, where we combine real and synthesized images, can significantly improve the classifier performance on small-scale datasets like Chaoyang, especially for underrepresented minority classes by up to 4-5\%. Our findings (results in supplementary) highlight the potential utility of synthesized images in improving the performance of downstream tasks.

\vspace{-4mm}
\section{Conclusion}

Our study proposes a novel approach that combines conditional DDIM with a vision transformer-based semantic encoder (ViT-DAE) for high-quality histopathology image synthesis. Our method outperforms recent GAN-based and vanilla DAE methods, demonstrating its effectiveness in generating more realistic and diverse histopathology images. The use of vision transformers in semantic encoder of DAE enables the holistic encoding of complex layouts specific to histo-pathology images, making it a promising solution for future research on image synthesis in digital pathology. 

%
%
\bibliographystyle{splncs04}
\bibliography{ref}

\begin{thebibliography}{10}
\providecommand{\url}[1]{\texttt{#1}}
\providecommand{\urlprefix}{URL }
\providecommand{\doi}[1]{https://doi.org/#1}

\bibitem{boyd2021self}
Boyd, J., Liashuha, M., Deutsch, E., Paragios, N., Christodoulidis, S.,
  Vakalopoulou, M.: Self-supervised representation learning using visual field
  expansion on digital pathology. In: Proceedings of the IEEE/CVF International
  Conference on Computer Vision (2021)

\bibitem{dino}
Caron, M., Touvron, H., Misra, I., J{\'e}gou, H., Mairal, J., Bojanowski, P.,
  Joulin, A.: Emerging properties in self-supervised vision transformers. In:
  Proceedings of the IEEE/CVF international conference on computer vision. pp.
  9650--9660 (2021)

\bibitem{dahan2022artifact}
Dahan, C., Christodoulidis, S., Vakalopoulou, M., Boyd, J.: Artifact removal in
  histopathology images. arXiv preprint arXiv:2211.16161  (2022)

\bibitem{dhariwal2021diffusion}
Dhariwal, P., Nichol, A.: Diffusion models beat gans on image synthesis.
  Advances in Neural Information Processing Systems  \textbf{34},  8780--8794
  (2021)

\bibitem{dosovitskiy2020image}
Dosovitskiy, A., Beyer, L., Kolesnikov, A., Weissenborn, D., Zhai, X.,
  Unterthiner, T., Dehghani, M., Minderer, M., Heigold, G., Gelly, S., et~al.:
  An image is worth 16x16 words: Transformers for image recognition at scale.
  arXiv preprint arXiv:2010.11929  (2020)

\bibitem{vqgan}
Esser, P., Rombach, R., Ommer, B.: Taming transformers for high-resolution
  image synthesis. In: Proceedings of the IEEE/CVF conference on computer
  vision and pattern recognition. pp. 12873--12883 (2021)

\bibitem{heusel2017gans}
Heusel, M., Ramsauer, H., Unterthiner, T., Nessler, B., Hochreiter, S.: Gans
  trained by a two time-scale update rule converge to a local nash equilibrium.
  Advances in neural information processing systems  \textbf{30} (2017)

\bibitem{ho2020denoising}
Ho, J., Jain, A., Abbeel, P.: Denoising diffusion probabilistic models.
  Advances in Neural Information Processing Systems  \textbf{33},  6840--6851
  (2020)

\bibitem{jose2021generative}
Jose, L., Liu, S., Russo, C., Nadort, A., Di~Ieva, A.: Generative adversarial
  networks in digital pathology and histopathological image processing: A
  review. Journal of Pathology Informatics  (2021)

\bibitem{kanwal2022devil}
Kanwal, N., P{\'e}rez-Bueno, F., Schmidt, A., Engan, K., Molina, R.: The devil
  is in the details: Whole slide image acquisition and processing for artifacts
  detection, color variation, and data augmentation: A review. IEEE Access
  (2022)

\bibitem{kather2018100}
Kather, J.N., Halama, N., Marx, A.: 100,000 histological images of human
  colorectal cancer and healthy tissue. Zenodo10  (2018)

\bibitem{kather2019deep}
Kather, J.N., Pearson, A.T., Halama, N., J{\"a}ger, D., Krause, J., Loosen,
  S.H., Marx, A., Boor, P., Tacke, F., Neumann, U.P., et~al.: Deep learning can
  predict microsatellite instability directly from histology in
  gastrointestinal cancer. Nature medicine  (2019)

\bibitem{kynkaanniemi2022role}
Kynk{\"a}{\"a}nniemi, T., Karras, T., Aittala, M., Aila, T., Lehtinen, J.: The
  role of imagenet classes in fr$\backslash$'echet inception distance. arXiv
  preprint arXiv:2203.06026  (2022)

\bibitem{kynkaanniemi2019improved}
Kynk{\"a}{\"a}nniemi, T., Karras, T., Laine, S., Lehtinen, J., Aila, T.:
  Improved precision and recall metric for assessing generative models.
  Advances in Neural Information Processing Systems  \textbf{32} (2019)

\bibitem{moghadam2023morphology}
Moghadam, P.A., Van~Dalen, S., Martin, K.C., Lennerz, J., Yip, S., Farahani,
  H., Bashashati, A.: A morphology focused diffusion probabilistic model for
  synthesis of histopathology images. In: Proceedings of the IEEE/CVF Winter
  Conference on Applications of Computer Vision. pp. 2000--2009 (2023)

\bibitem{naseer2021intriguing}
Naseer, M.M., Ranasinghe, K., Khan, S.H., Hayat, M., Shahbaz~Khan, F., Yang,
  M.H.: Intriguing properties of vision transformers. Advances in Neural
  Information Processing Systems  \textbf{34},  23296--23308 (2021)

\bibitem{dae}
Preechakul, K., Chatthee, N., Wizadwongsa, S., Suwajanakorn, S.: Diffusion
  autoencoders: Toward a meaningful and decodable representation. In:
  Proceedings of the IEEE/CVF Conference on Computer Vision and Pattern
  Recognition. pp. 10619--10629 (2022)

\bibitem{rombach2022high}
Rombach, R., Blattmann, A., Lorenz, D., Esser, P., Ommer, B.: High-resolution
  image synthesis with latent diffusion models. In: Proceedings of the IEEE/CVF
  Conference on Computer Vision and Pattern Recognition. pp. 10684--10695
  (2022)

\bibitem{ronneberger2015u}
Ronneberger, O., Fischer, P., Brox, T.: U-net: Convolutional networks for
  biomedical image segmentation. In: Medical Image Computing and
  Computer-Assisted Intervention--MICCAI 2015: 18th International Conference,
  Munich, Germany, October 5-9, 2015, Proceedings, Part III 18. pp. 234--241.
  Springer (2015)

\bibitem{runz2021normalization}
Runz, M., Rusche, D., Schmidt, S., Weihrauch, M.R., Hesser, J., Weis, C.A.:
  Normalization of he-stained histological images using cycle consistent
  generative adversarial networks. Diagnostic Pathology  (2021)

\bibitem{Seitzer2020FID}
Seitzer, M.: {pytorch-fid: FID Score for PyTorch}.
  \url{https://github.com/mseitzer/pytorch-fid} (August 2020), version 0.3.0

\bibitem{song2020denoising}
Song, J., Meng, C., Ermon, S.: Denoising diffusion implicit models. arXiv
  preprint arXiv:2010.02502  (2020)

\bibitem{song2020score}
Song, Y., Sohl-Dickstein, J., Kingma, D.P., Kumar, A., Ermon, S., Poole, B.:
  Score-based generative modeling through stochastic differential equations.
  arXiv preprint arXiv:2011.13456  (2020)

\bibitem{vaswani2017attention}
Vaswani, A., Shazeer, N., Parmar, N., Uszkoreit, J., Jones, L., Gomez, A.N.,
  Kaiser, {\L}., Polosukhin, I.: Attention is all you need. Advances in neural
  information processing systems  \textbf{30} (2017)

\bibitem{veeling2018rotation}
Veeling, B.S., Linmans, J., Winkens, J., Cohen, T., Welling, M.: Rotation
  equivariant cnns for digital pathology. In: Medical Image Computing and
  Computer Assisted Intervention--MICCAI 2018: 21st International Conference,
  Granada, Spain, September 16-20, 2018, Proceedings, Part II 11. pp. 210--218.
  Springer (2018)

\bibitem{wei2019generative}
Wei, J., Suriawinata, A., Vaickus, L., Ren, B., Liu, X., Wei, J., Hassanpour,
  S.: Generative image translation for data augmentation in colorectal
  histopathology images. Proceedings of machine learning research
  \textbf{116}, ~10 (2019)

\bibitem{xiao2021tackling}
Xiao, Z., Kreis, K., Vahdat, A.: Tackling the generative learning trilemma with
  denoising diffusion gans. arXiv preprint arXiv:2112.07804  (2021)

\bibitem{xue2021selective}
Xue, Y., Ye, J., Zhou, Q., Long, L.R., Antani, S., Xue, Z., Cornwell, C.,
  Zaino, R., Cheng, K.C., Huang, X.: Selective synthetic augmentation with
  histogan for improved histopathology image classification. Medical image
  analysis  \textbf{67},  101816 (2021)

\bibitem{zhu2021hard}
Zhu, C., Chen, W., Peng, T., Wang, Y., Jin, M.: Hard sample aware noise robust
  learning for histopathology image classification. IEEE Transactions on
  Medical Imaging  (2021)

\end{thebibliography}
%





\end{document}


\begin{large}
\centerline{\textbf{Supplementary Material}}
\end{large}
%
%
%

%
%

%
%
%


\begin{table}
\small
\centering
\caption{\textbf{Proof-of-concept downstream analysis on the Chaoyang dataset.} We leveraged a ViT pretrained on ImageNet for classification on the Chaoyang dataset by utilizing synthesized images generated from our proposed method. To enhance the classifier's performance, we incorporated both real ($\mathcal{R}$) and synthesized ($\mathcal{S}$) images, and observed that the inclusion of synthesized images resulted in performance improvement. We found that this hybrid training approach was particularly beneficial for improving the classification (F1-score in $5\%$) of underrepresented minority classes, such as adenoma. $\#$ samples for normal, serrated, adenocarcinoma, adenoma = 1816, 1163, 2244, and 937, respectively.}\label{downstream}
\begin{tabular}{|c|c|c|c|c|c|}
\hline
 Data & Acc & $F1_{normal}$ & $F1_{serrated}$ & $F1_{adenocarc}$ & $F1_{adenoma}$ \\\hline
  $\mathcal{R}$&  78.40±0.28 & 78.48±0.26 & 47.00±1.60 & 92.26±0.26 & 68.30±0.69\\\hline
 $\mathcal{S}$ &  75.48±0.86 & 72.95±2.41
& 46.86±1.72 & 91.34±0.39 & 67.73±0.86\\\hline
 $\mathcal{R}$+$\mathcal{S}$ & \textbf{80.12±0.33} & \textbf{78.60±0.94}
& \textbf{51.79±2.18} & \textbf{93.50±0.48} & \textbf{73.74±1.86}\\\hline
\end{tabular}
\end{table}
\begin{table}
\small
\centering
\caption{\textbf{Class-wise FID ($\downarrow$) scores on the NCT-CRC dataset.} To generate class-specific images, we trained independent latent DDIM models on semantic distributions for each class. We compared the synthesized images (5k/class) with the real images from the training set and calculated the FID score. Our model achieved the lowest FID scores for most of the classes (8/9 classes); for the ADI class, our result was comparable to VQ-GAN. These findings suggest that our proposed ViT-DAE model effectively captures high-level semantic information and improves the overall quality of the generated images.}\label{fidclscrc}
\begin{tabular}{|l|c|c|c|c|c|c|c|c|c|}
\hline
Methods &  ADI & BACK & DEB & LYM & MUC & MUS & NORM & STR & TUM\\
\hline
VQ-GAN &  \textbf{27.73} &58.93 & 32.41 &30.24 & 43.77 &32.62 &62.73 &37.50 &43.50\\
DAE &  31.48 &30.31 & 15.31 & 21.61 & 24.47 & 26.99 & 22.76 & 13.94 & 20.92\\\hline
ViT-DAE (Ours) & 27.85 & \textbf{28.55} & \textbf{12.72} & \textbf{15.28} & \textbf{16.70} & \textbf{24.19}& \textbf{16.06} & \textbf{11.70} & \textbf{15.29}\\
\hline
\end{tabular}
\end{table}
\begin{figure}[t]
\begin{center}
\includegraphics[width=0.8\linewidth]{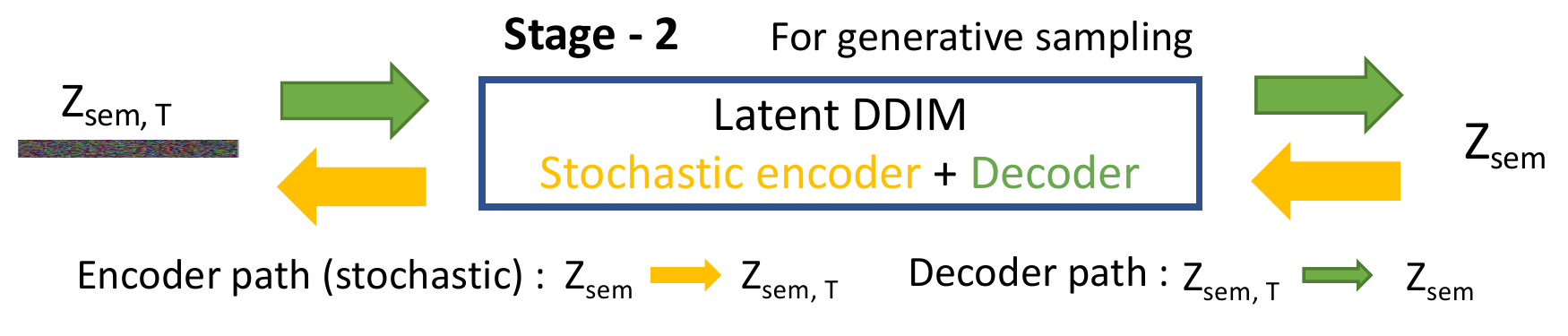}
\end{center}
\caption{Generative sampling process. In Stage-2, a latent DDIM is trained to learn the distribution of semantic representations of data.  }\label{latentDDIM}
\end{figure}

\begin{figure}[h]
\begin{center}
\includegraphics[width=.77\linewidth]{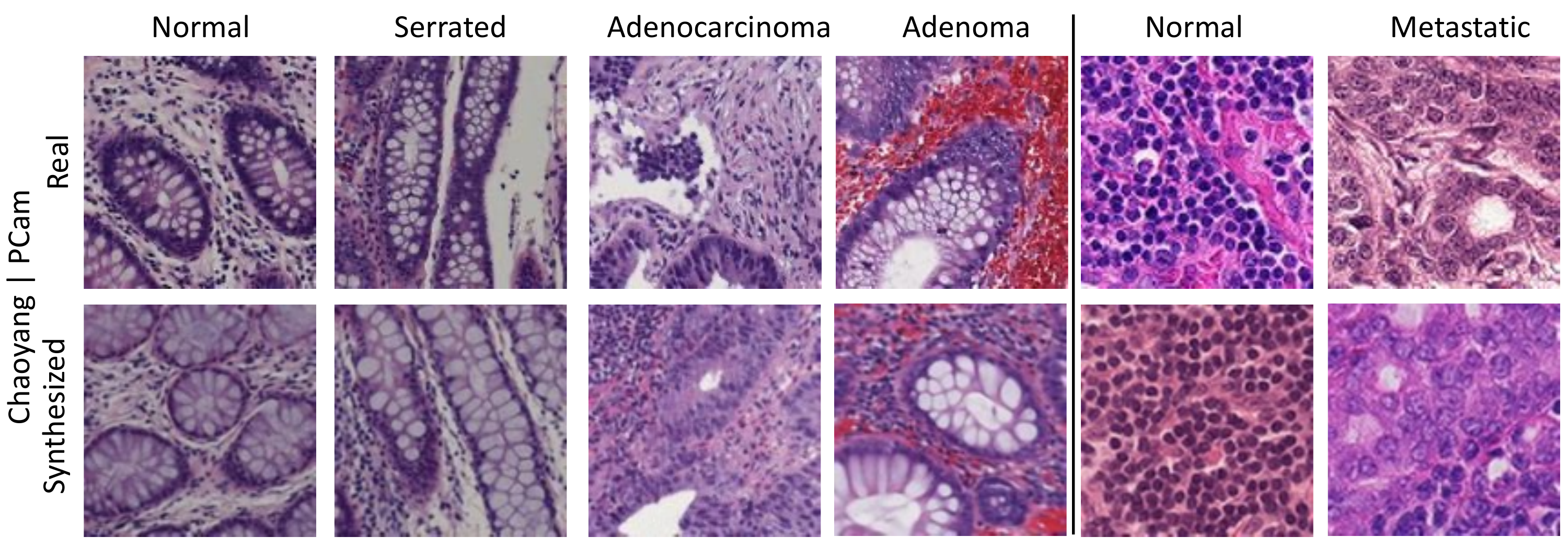}
\end{center}
\caption{Real and synthesized images using ViT-DAE on Pcam and Chaoyang} \label{visualization}
\end{figure}

\begin{figure}[h]
 \centering
\includegraphics[width=0.8\textwidth]{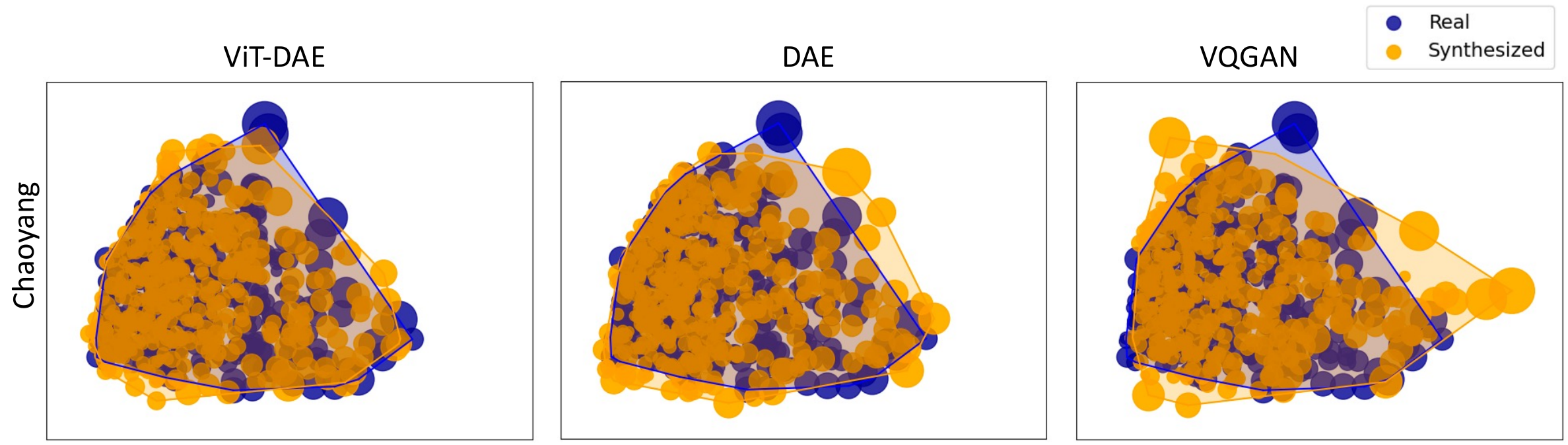}

\caption{Manifold visualization for Chaoyang. DAE and ViT-DAE generate distributions that are very comparable to real images while outperforming the generated manifold from VQ-GAN. } 
\label{manifold}
\end{figure}


\begin{figure}[h]
\begin{center}
\includegraphics[width=.9\linewidth]{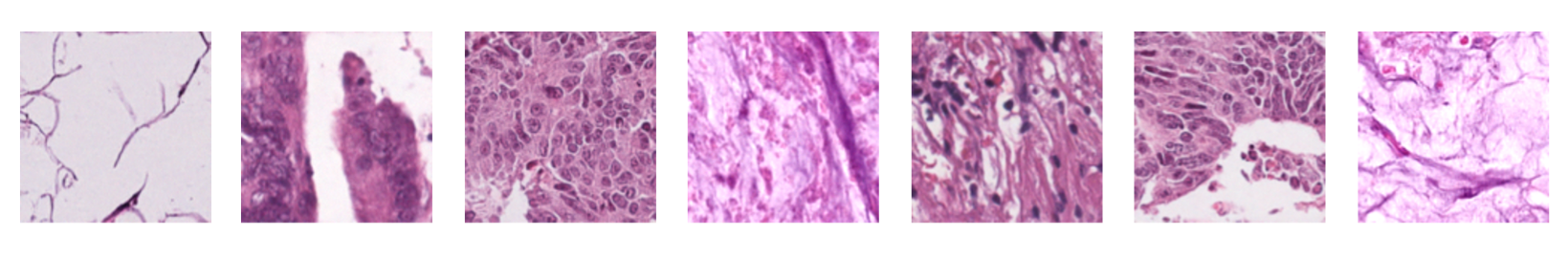}
\end{center}
\caption{\textbf{Synthesized images using DDPM on NCT-CRC.} We leverage DDPM[15] to synthesize images to compare with DDIM. A fast FID score (20k synthesized images) is calculated primarily due to high sampling time of DDPM. $FID_{DDPM,20k}=30.84, FID_{ViT-DAE,20k}=12.55$. This shows that our conditional DDIM exhibits superior image generation capabilities in terms of producing high-quality images compared to DDPM.} \label{ddpm}
\end{figure}

\bibliographystyle{splncs04}



